\documentclass[review]{elsarticle}

\usepackage{lineno,hyperref}

\usepackage{times}
\usepackage{url}
\usepackage{latexsym}
\usepackage{longtable}
\usepackage{mathrsfs}

\usepackage{amssymb}
\usepackage{amsmath}
\usepackage{multicol}
\usepackage{multirow}
\usepackage{empheq}
\usepackage{float}

\usepackage{caption}
\usepackage[marginal]{footmisc}

\usepackage{amssymb}
\usepackage{array,graphicx,subfigure,wrapfig}

\usepackage{amsmath,amssymb}
\usepackage{leftidx}

\usepackage{subfloat}
\usepackage{threeparttable}
\usepackage{makecell}
\usepackage{CJKutf8}

\usepackage{latexsym}
\usepackage{algorithm}
\usepackage[noend]{algpseudocode}
\usepackage{bm}

\floatstyle{plain}
\newfloat{myalgo}{tbhp}{mya}
\newenvironment{Algorithm}[2][tbh]%
{\begin{myalgo}[#1]
\centering
\begin{minipage}{#2}
\begin{algorithm}[H]}%
{\end{algorithm}
\end{minipage}
\end{myalgo}}

\usepackage{lineno}

\usepackage[figuresright]{rotating}

\usepackage{lineno}

\usepackage[figuresright]{rotating}

\modulolinenumbers[5]

\begin{document}

\begin{frontmatter}

\title{Stroke Sequence-Dependent Deep Convolutional Neural Network for Online Handwritten Chinese Character Recognition}

\author[addr]{Baotian Hu}
\ead{baotianchina@hitsz.edu.cn}

\author[addr]{Xin Liu}
\ead{hit.liuxin@gmail.com}

\author[addr]{Xiangping Wu}
\ead{wxpleduole@gamil.com}

\author[addr]{Qingcai Chen\corref{maincorrespondingauthor}}
\ead{qingcai.chen@hitsz.edu.cn}

\cortext[maincorrespondingauthor]{Corresponding author}
\address[addr]{Key Laboratory of Network Oriented Intelligent Computation\\  Harbin Institute of Technology Shenzhen Graduate School, Shenzhen, PR China}

\begin{abstract}

In this paper, we propose a novel model, named Stroke Sequence-dependent Deep Convolutional Neural Network (SSDCNN), using the stroke sequence information and eight-directional features for Online Handwritten Chinese Character Recognition (OLHCCR). On one hand, SSDCNN can learn the representation of Online Handwritten Chinese Character (OLHCC) by incorporating the natural sequence information of the strokes. On the other hand, SSDCNN can incorporate eight-directional features in a natural way. Firstly, SSDCNN takes the stroke sequence as input by transforming them into stacks of feature maps according to their writing order. And then, the fixed length stroke sequence dependent representations of OLHCC are derived via a series of convolution and max-pooling operations. Thirdly, stroke sequence dependent representation is combined with the eight-directional features via a number of fully connected neural network layers. Finally, the softmax classifier is used as recognizer. In order to train SSDCNN, we divide the process of training into two stages: 1) The training data is used to pre-train the whole architecture until the performance tends to converge. 2) Fully-connected neural network which is used to combine the stroke sequence-dependent representation with eight-directional features and softmax layer are further trained. Experiments were conducted on the OLHCCR competition tasks of ICDAR 2013. Results show that, SSDCNN can reduce the recognition error by 50\% (5.13\% vs 2.56\%) compared to the model which only use eight-directional features. The proposed SSDCNN achieves 97.44\% accuracy which reduces the recognition error by about 1.9\% compared with the best submitted system on ICDAR2013 competition. These results indicate that SSDCNN can exploit the stroke sequence information to learn high-quality representation of OLHCC. It also shows that the learnt representation and the classical eight-directional features complement each other within the SSDCNN architecture.
\end{abstract}


\end{frontmatter}

\section{Introduction}
\label{intro}

\begin{figure}
  \centering
  \subfigure[]{
    \includegraphics[width=.50\textwidth]{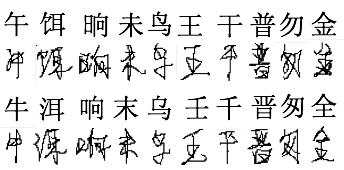}
    }
  \subfigure[]{
    \includegraphics[width=.36\textwidth]{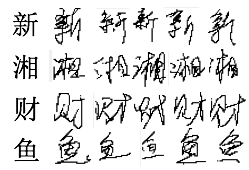}
   }
  \caption{(a) shows some the similar Chinese characters written by one writers. The printed row lists the gold characters.  (b) depicts some Chinese characters written by different writers. The left column lists the gold characters}
  \label{fig1}
\end{figure}

Handwritten Character Recognition (HCR) is the process of mapping the handwritten characters to input text of machines automatically. HCR can benefit to the interaction between human and machine in a lot of applications, such as handwriting input for smart phone, postal mail sorting, bank check processing, form processing and so on ~\cite{hir2008,liu2006}. Conventional HCR includes offline and online HCR~\cite{online_survey2000}. The major difference between them is the representation of data~\cite{liu2013}. The data for offline HCR is usually represented as static image. While online handwritten  characters are composed of strokes represented by the pen-tip movement traces with the pen-up/pen-down switching. Here a stroke is defined as the sequence of points sampled between consecutive pen-down and pen-up transitions~\cite{connell1998}. This work is focused on the Online Handwritten Chinese character Recognition (OLHCCR). Compared with the performance of offline handwritten Chinese character recognition (OHCCR), for the benefits of stroke information,the OLHCCR reaches higher reported performance. For example, the ICDAR2013 handwritten Chinese Character recognition competition tasks report the best results of 94.77 and 97.39 for OHCCR and OLHCCR respectively~\cite{liu2013}. Despite the tremendous works during the past decades, OLHCCR remains a major challenge for three reasons:  1) large number of character classes, the total number of Chinese characters is about 30,000 and the daily used ones are more than 5000; 2) many Chinese characters are quit similar, some examples shown in Fig.~\ref{fig1} (a); 3) the variability of writing styles for each character is very high, some examples shown in Fig.~\ref{fig1} (b).

To address these issues, conventional methods for OLHCCR usually consist of three stages, i. e., data preprocessing, feature extraction and classification, as shown in Fig.~\ref{fig2}. The purpose of data preprocessing is to address the high variability issue of handwritten characters by using deformation or normalization techniques~\cite{liu2013}. It is important for improving OLHCCR since the handwritten samples collected is still not enough. Feature extraction methods are the most important techniques for conventional OLHCCR models. Some models use corresponding static image as one part of input for OLHCCR, which composes one important type of OLHCCR features. In addition to the image, there are still stroke features, such as the path signature feature~\cite{cnn2013}, and the widely used eight-directional features~\cite{liu2006} etc., have been used for OLHCCR. Considering about the large number of classes for OLHCCR, constructing a proper classifier is also a very challenging task. Among published conventional classifiers for OLHCCR, the modified quadratic discriminant function (MQDF)~\cite{mqdf} and discriminative quadratic discriminant function (DLQDF)~\cite{zhou2016} had played very important roles. To get higher recognition precision, multiple types of data preprocessing methods, normalization methods, feature extraction methods and classifiers usually need to be combined together. Liu et al had given a detail introduction and comparison of main conventional OLHCCR techniques that covered above stages~\cite{liu2013}. On their CASIA-OLHWDB (1.0 and 1.1), they get the best performance with a relatively complicate model that integrated pseudo 2D character normalization, normalization-cooperated gradient/trajectory direction feature extraction and discriminative feature extraction (DFE) and DLQDF. It is obvious that to improve the performance of OLHCCR through conventional techniques has become really a challenging task.

In recent year, deep convolutional neural network(DCNN) has shown superior performance on many tasks such as image classification~\cite{cnn2012} and speech recognition~\cite{cnn_speech}. DCNN can learn local representation of image through number of convolutional layers operations on the small patches of the image. At the same time, The layer-by-layer max pooling on the convolutional representations allows it to acquire high level representation of the image (such as the stroke level and character level features). The mechanism of the DCNN makes it invariant to rotation, translation, scaling. For OLHCCR, Ciresan et.al~\cite{cnn2012} treats the handwritten sample simply as an image bitmap. Although this model beats the best result of the traditional models, it ignores the important stroke information such as the stroke sequence and stroke direction which are important to OLHCCR. Graham submitted a DCNN-based model for ICDAR2013 Online Isolated Chinese Character recognition competition tasks~\cite{cnn2013,yinfei2013}. This model incorporates the stroke direction feature and the curvature of the stroke by computing their path signature. The signature is represented by six input feature matrices that are corresponding to displacement and curvature of strokes~\cite{yinfei2013}, which are then jointed with the matrix of the static image generated from OLHCC as the input of an sparse adapted DCNN architecture. To improve the classification performance, the preprocessing of Gaussian blur and elastic distortions are also used as data preprocessing~\cite{cnn2013}. This model wins the competition, which indicates that DCNN have great potential to learn the important information by considering the characteristics of online handwritten characters. Yang et.al~\cite{cnn2015} further propose to use various domain-specific knowledge i.e, deformation transformation, non-linear normalization, imaginary stroke, path signature features, eight-directional features, to improve DCNN based OLHCCR. Their method mainly depends on the budget for network scale and time consumption, where the property of OLHCC is not considered well.

Instead of integrating more handcraft features into DCNN model to improve the performance of OLHCCR, in this paper, we propose a more feasible architecture that only incorporate the stroke sequence-dependent representation and traditional eight-directional features of OLHCC. It is called stroke sequence-dependent deep convolutional neural network (SSDCNN) model. Firstly, DCNN is used to learn the hight-quality representation of OLHCCs by directly exploiting the stroke sequence features using the original image of each stroke as one input matrix. The image is generated by a very simple way, i.e., by drawing lines for each two trajectory points of the stroke and scaling them into a fixed dimensions bitmap. And then, we combine the representation derived from DCNN with the traditional eight-directional features. In order to train SSDCNN efficiently, we design a two stages algorithm to train SSDCNN. Firstly, we use the data to pre-train the whole framework. In order to merge the two representation efficiently, we further train the fullly-connected multiple layer perceptron with a softmax layer by fixing the parameter of the DCNN. Experiment results on the competitive ICDAR2013 Online Isolated Chinese Character recognition task show that the proposed SSDCNN achieves the accuracy of 97.44\% which is the state of the art result within this task context and surpass the ability of human. The main contributions of this paper include: 1) We propose a very effective model of learning stroke sequence-dependent representation using the DCNN for OLHCCR. 2) We propose a simple and feasible architecture that effectively integrates two complementary type of features: the learnt stroke sequence-dependent representation and the traditional eight-directional features. 3) We achieve the best performance of 97.44\% on the same context of ICDAR2013 Online Isolated Chinese Character Recognition task.

The remainder of this  paper is organized as follows, In section 2, we briefly review some related works on OLHCCR. Section 3 describes the proposed SSDCNN. Section 4 introduces how to train the whole framework. Section 5 presents experimental results and Section 6 concludes the paper.


\section{Related Work}

\begin{figure}[!h]
\centerline{\includegraphics[width=1.0\textwidth]{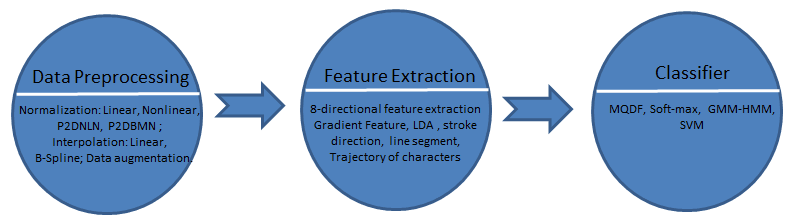}}
\caption{The flowchart of the traditional methods.}
\label{fig2}
\end{figure}

Intensive research efforts have been made on OLHCCR since 1960s. Generally, the proposed methods can be divided into structural methods and statistical methods~\cite{liu2004_2}: Before 1990s, most of works are related to structural methods, which is more relevant to human learning and perception. Afterward, statistical methods can achieve higher recognition accuracy by learning from samples. However both structural methods and statistical methods are based on handcraft features. We first review some related works on online isolated character recognition according to the flowchart Fig.~\ref{fig2}. And then, we will describe some recent works related to DCNN based method on OLHCCR.

For OLHCCR, data preprocessing is to normalize the size variations and remove data noises that may lead to the decrease of the performance~\cite{Husain2007}. Data Normalization and interpolation are the frequently used data processing methods. Data Normalization is to transform strokes of OLHCC to a fixed size so as to extract features with common dimensionality~\cite{liu2005}. How to restore the deformation of handwritten characters shape variation within class is the main interest of researchers. Linear normalization is not sufficient to meet deformation restoration requirement~\cite{liu2005}. Many methods such as Line Density-based Normalization, Curve-fitting-based normalization and Pseudo 2D Normalization are proposed to replace linear normalization. Unlike alphanumeric characters, most of Chinese characters are composed of multiple strokes. The nonlinear normalization(NLN) method has the ability to regulate the stroke spacing~\cite{lee1994}. The NLN based on line density equalization has been proven very efficient~\cite{lee1994}. In order to use line density-based method to OLHCC,  Phan et.al~\cite{phan2011} convert an OLHCC to a 2D image. For Curve-fitting-based normalization methods, three algorithm are often used such as bi-moment normalization(BMN)~\cite{liu2003}, centroid-boundary alignment (CBA), and modified CBA (MCBA)~\cite{liu2004}. They can be used to OLHCC after slightly modified~\cite{liu2006,phan2011}. Pseudo two-dimensional normalization is popularly used for off-line HCCR~\cite{liu2005}. Horiuchi et.al proposed a Pseudo two-dimensional nonlinear shape normalization methods(P2DNLN) and popularly used ~\cite{hor1997} for offline HCCR. However, the time complexity of P2DNLN is very high. There are various pseudo 2D normalization methods such as pseudo 2D moment normalization (P2DMN), pseudo 2D bi-moment normalization (P2DBMN)~\cite{hailong2005}, and pseudo 2D MCBA (P2DCBA). Liu et.al~\cite{liu2006} adapt these methods to online Japanese Character Recognition~\cite{liu2006}. High speed of handwriting may lead to missing points for OHCC. The solution of this  problem is the interpolation~\cite{spine2004}. The frequently used method is linear interpolation~\cite{linear}, Bezier interpolation and B-spline interpolation~\cite{bspline}.

The performance of OLHCCR system depends largely on the feature extraction, since the feature vectors are the direct input to the classifier. For handwritten Chinese Character recognition, directional features~\cite{liu2013} and gradient features~\cite{yinfei2013} are usually used. Kawamura et.al~\cite{Kawamura1992} propose to use 4-directional features for online Handwritten Japanese recognition. Bai and Huo extends 4-directional features to eight-directional features for OLHCCR. Ding et.al~\cite{ding2009} extract the 8-diretional features on imaginary strokes for OLHCCR and Liu et.al~\cite{liu2006} extract the original stroke direction and normalized direction features for OLHCCR. Yun et.al~\cite{histogram2008} propose to use the combination of distance map and direction histogram to recognize online handwritten digits and freehand sketches. Zhu et.al, participate to the ICDAR 2013 Chinese Handwriting Recognition Competition\footnote{http://www.nlpr.ia.ac.cn/events/CHRcompetition2013/competition/Home.html} and they submit the system named TUAT by using the histograms of normalized stroke direction ~\cite{yinfei2013}. TUAT achieve 93.85\% ranking eighth in the ninth systems. Su et.al submit a system named HIT by using gradient feature to ICDAR2013 competition which achieve 95.18\% ranking fourth in the ninth systems~\cite{yinfei2013}.

The classifier is also very important to OLHCCR, since the classifier is the output layer for specific class of OLHCC. Various classifiers can be used for recognition, such as statistical classifiers and support vector machines(SVM)~\cite{svm}. Modified quadratic discriminant function(MQDF) proposed by Kimura et.al ~\cite{mqdf} is a compact Gaussian classifier. Compared to quadratic discriminant function (QDF) MQDF needs lower storage and time complexity. MQDF is the generative model and its parameter can be derived from  minimum classification error(MCE-MQDF)~\cite{mce, hailong2005} and perceptron learning(PL-MQDF)~\cite{plmqdf}. Discriminative learning quadratic discriminant function (DLQDF) is actually the version of MQDF where the parameters are optimized by discriminative training under the minimum classification error(MCE) criterion~\cite{zhou2016}. DLQDF is popularly used for handwriting Chinese character recognition~\cite{liu2013,zhou2016}. Nearest prototype classifier(NPC) learns a set of points, named prototypes, from training data feature space. The test instances are classified to the class of nearest prototype~\cite{npc}. There are various methods to learn the prototype of NPC such as clustering, learning vector quantization(LVQ)~\cite{lvq}. NPC is also often used for OLHCCR~\cite{liu2013, zhou2016,plmqdf}. In order to reduce the time complexity of recognition engines, two strategies are often used: 1) use Fisher linear discriminant analysis (FLDA) to reduce the dimension of the features~\cite{liu2005,liu2006,plmqdf,tao2014}; 2) use hierarchical classifiers instead of single classifier~\cite{tao2014,plmqdf}. In another word, feed the representation into a first level classifier for coarse classification which get a small number of candidates, and then the second level classifier to get the final category of the input by taking the input of the representations. For example, various of works use the Euclidean distance to class means for selecting the top $k$ candidates and use the MQDF to get the final labels from these $k$ candidates~\cite{liu2006,liu2013}.

Our work is highly related to the line of research on the deep neural network models. In recent years,  deep neural networks have shown its noteworthy success in computer vision~\cite{cnn2012} and speech recognition~\cite{cnn_speech}. Ciresan et.al~\cite{cnn2012} proposed the multi-column deep neural network(MCDNN) for image classification. The MCDNN works on raw pixel intensities directly and learns the feature representation in supervised way~\cite{cnn2012}. MCDNN beats other methods by a large margin and reaches close to the human performance. The coordinate of each online character is drawn on a fixed size 40x40 box in the center of 48x48 image, which actually transform the online character into image bitmap. This method lost much important information of online characters, such as the point direction, stroke sequence. Motivated by the ``signature'' from theory of differential equations, Graham~\cite{cnn2013} extract path signature feature from online characters. They stack the path features to get a $N\times N\times M$ representation of the online character according to the order of  iterated integral, where the zeroth, first and second iterated integrals correspond to the information of off-line version of the character, the point direction of the stroke and the curvature of the stroke. They feed the $N\times N\times M$  representation into a large DCNN. They test their method on the ICDAR2013 Online Isolated Chinese Character recognition competition and won the first place. However, the input of the DCNN is relatively complex and the model also ignores the sequence information of the strokes. Yang et.al~\cite{cnn2015} propose to use various domain-specific knowledge i.e, deformation transformation, non-linear normalization,imaginary stroke, path signature features, eight-directional features, to improve DCNN based OLHCCR.  They use hybrid serial-parallel strategy to combine the various domain-specific knowledge DCNNs.


\section{Stroke Sequence-dependent Deep Convolutional Neural Network}

We design a novel model to learn the stroke sequence-dependent representation of OLHCCs and combine the traditional eight-directional features, which is named SSDCNN.  The overall framework of SSDCNN is depicted as Fig.~\ref{model}. The characteristics of this model can be summarized as follows:

1) The DCNN is used to learn the high-quality representation of OLHCCs, which is stroke sequence-dependent and can learn the stroke sequence information and structural shape of OLHCCs from large scale instances.

2) The statistical feature is integrated into SSDCNN via a deep fully-connected neural network. Specifically, the eight-directional features are used, which has been proven efficient in OLHCCR. Hence, SSDCNN can preserve the strength of statistical features.

\subsection{Learn Stroke Sequence-dependent Representation via DCNN}

In contrast to English words which are composed of one or more alphabet letters, most of Chinese characters are made up of one or more basic components namely stroke. The eight basic strokes are vertical, horizontal, angular, slash, saber, hooked, tick and dot as illustrated in Fig.~\ref{stroke} (a). The basic strokes and strokes order of one character are very important to determine what character the OLHCC is. Traditionally, when the teacher teaches the child to write Chinese characters, they will start from basic strokes. And then, the child will learn to write characters by obeying the standard order of the basic strokes. There are some basic rules to guide the child to write the character such as ``From top to bottom'', ``From left to right'', ``Horizontal before vertical'' and ``Outside before inside''. Fig.~\ref{stroke} (b) shows some characters with correct stroke order according to ``Modern Chinese Commonly Used Character Stroke Order Standard'' published in 1997. Writing Chinese characters according to the correct stroke order can greatly facilitate learning and memorization. Although writers will develop their individuality instead of observing the standard sequence of strokes strictly, some basic rules still govern them to write characters. Li et.al studied 61 commonly encountered Chinese characters by inviting 372 persons to write them, the majority (around 60\%) of them write the character according to the correct sequence of strokes~\cite{li2007}. Hence, The sequence of strokes is an very important feature for specific Chinese character. We propose to use DCNN to learn the stroke sequence-dependent representation.

\begin{figure}[!h]
\centerline{\includegraphics[width=1.\textwidth]{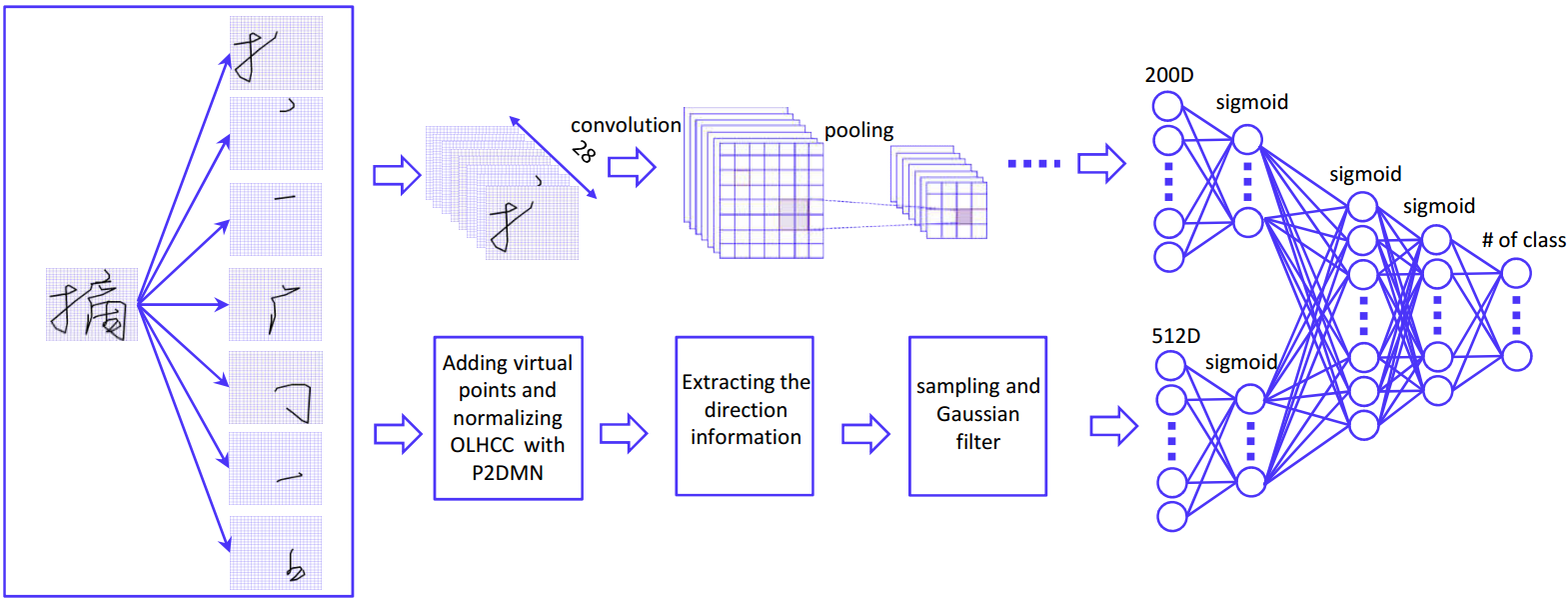}}
\caption{The overall framework of SSDCNN. }
\label{model}
\end{figure}

A OLHCC is represented by a sequence of strokes which are represented by sequence of coordinate points. Given a OLHCC $\mathscr{C}$ .
\begin{equation}
\mathscr{C}=({({x_1}^1,{y_1}^1),...({x_1}^i,{y_1}^i)},...,{({x_m}^1,{y_m}^1),...,({x_m}^j,{y_m}^j)})
\label{eq}
\end{equation}

\noindent Where $({x_n}^l,{y_n}^l)$ is the $l^{th}$ coordinate point of the $n^{th}$ stroke. We firstly project strokes onto isolated fixed-size feature maps. The values of positions where the stroke passes are set to ``1'', otherwise ``0''. We get the sequence of stroke feature maps $\mathscr{F}$.

\begin{equation}
 \mathscr{F}=(s_1,...,s_m)
\label{eq111}
\end{equation}

\noindent Where $s_i$ is the $i^{th}$ stroke feature map of Character $C$, which is a fixed size image.

For the input of the DCNN, we stack the stroke feature maps $\mathscr{F}$ according to the order of strokes shown as Fig.~\ref{model}. The variable number of strokes for different OLHCC can be handled by putting all-zero feature maps after the last stroke map $s_m$ until the maximum length. In this way, the stroke sequence information is incorporated into the model naturally. And then we do the convolution operation on the input feature maps.

For the Layer-1, we use convolution operation on the input of the stroke maps, we get output as follows:

\begin{equation}
z^{(1, f)}_{i,j} =  g(\hat{\mathbf{z}}^{(0)}_{i,j})\cdot \sigma(\mathbf{W}^{(1,f)} \hat{\mathbf{z}}^{(0)}_{i,j} + b^{(1,f)}).
\end{equation}

\begin{itemize}
  \item $z^{(1,f)}_{i}(\mathbf{x})$ gives the output of feature map of type-$f$  for location $i$ in Layer-$1$;
   \item $w^{(1, f)}$ is the parameters for $f$ on Layer-$1$, with matrix form $W^{(1)} \overset{\text{def}}{=} [w^{(1,1)},\\ \cdots,w^{(1,F_{1})}] $;
   \item $\sigma(\cdot)$ is the activation function (i.e.,Relu~\cite{relu})
  \item $\hat{\mathbf{z}}^{0}_{i}$ denotes the concatenation of the values in the convolution windows of input layer of the OLHCC for the convolution at location $i$.
\end{itemize}

In this way, the local feature of each stroke and the stroke sequence information will be learned. After the first convolution layer, we perform a max-pooling operation on non-overlapping $2\times 2$ windows.

\begin{equation}
z_{i,j}^{(2,f)} = \max(\{z_{2i-1,2j-1}^{(1,f)}, z_{2i-1,2j}^{(1,f)},z_{2i,2j-1}^{(1,f)},z_{2i,2j}^{(1,f)}\}).
\end{equation}
In Layer-3, we perform a 2D convolution on $k_3\times k_3$ windows of output from Layer-2:
\begin{equation}
z^{(3, f)}_{i,j} =  g(\hat{\mathbf{z}}^{(2)}_{i,j})\cdot \sigma(\mathbf{W}^{(3,f)} \hat{\mathbf{z}}^{(2)}_{i,j} + b^{(3,f)}).
\end{equation}
This could go on for more layers of convolution and max-pooling until the fixed length representation vector. Finally we get the representation $\mathscr{R} = (r_1,r_2,...,r_n)$, where $n$ is the dimension of the representation and no longer related to the stroke number of the given OLHCC $\mathscr{C}$.

\begin{figure}
  \centering
  \subfigure[Eight basic Chinese Strokes]{
    \includegraphics[width=.50\textwidth]{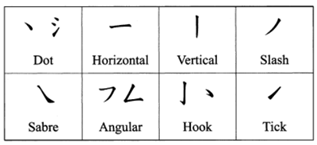}
    }
  \subfigure[Stroke Order of Four Chinese Characters]{
    \includegraphics[width=.37\textwidth]{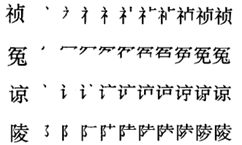}
   }
  \caption{(a) is the eight basic Chinese strokes, (b) depicts four Chinese characters with standard sequence of strokes. The sequence of the strokes is adopted from ``Modern Chinese Commonly Used Character Stroke Order Standard'' published by China National Language And Character Working Committee and General Administration of Press and Publication of the Peoples' Republic of China in 1997.}
  \label{stroke}
\end{figure}

\subsection{Integrate Eight-directional Features into SSDCNN}
The designed DCNN is to learn the stroke sequence-dependent representation of the OLHCC. Although it has the potential to learn the structure features of OLHCC theoretically by using the local convolution and max pooling operations, it requires enough label data to train because of the writing variation of OLHCC. The conventional handcraft features which is designed by extensively analyzing the problem and need not to be trained are useful. The eight-diretional feature has been proven efficient in the long term period. The eight-directional features can represent stroke shape and direction by mathematically analyzing point directions to its surrounding points. The weakness of the eight-directional features is that it can not represent the directional information and positional relationship of intra-strokes points, because it only extracts the direction feature between the ending point of one stroke and the starting point in its following stroke regardless of the positional relationship between other points in different strokes. The stroke sequence-dependent and the traditional eight-directional features can compensate with each other by designing a reasonable way of combing them together. In this part, we combine the two different features by using multiple layer perceptron. Because, the eight-directional features and stroke sequence-dependent representation come from different types of methods, the scale of the two types features are different. The element value of eight-direction feature range from 0 to 1. Once generated, it can never be modified. Since we choose ReLU as the activation function in convolutional layers, the stroke sequence-dependent representation may be unrestricted numerical values ranging from 0 to infinite. If we directly concatenate two features as one, the different scale of them may cause data inconsistency. In order to solve the problem, in our model, we feed the stroke sequence-dependent representation and eight-directional features into two different fully-connected layer with activation function Sigmoid as Eq.~\ref{sigmoid}.
\begin{equation}
\tilde{h} \leftarrow \frac{1}{1+e^h}
\label{sigmoid}
\end{equation}
Given the representation $\mathscr{L}=(l_1,l_2,...,l_{512})$ gotten by using eight-directional features extracting method and $\mathscr{R}$ which is learned by using DCNN. We can get the $\tilde{\mathscr{L}}$ and $\tilde{\mathscr{R}}$ by different fully-connected layers respectively.
\begin{equation}
\tilde{\mathscr{L}} =  \sigma(\mathbf{{W_l}} \mathscr{L}^T + b_l)
\label{Eq1}
\end{equation}

\begin{equation}
\tilde{\mathscr{R}} =  \sigma(\mathbf{{{W_r}}} \mathscr{R}^T + b_r)
\label{Eq2}
\end{equation}

\noindent Where $W_l\in\mathbb{R}^{n\times n}$, $b_o\in\mathbb{R}^{n\times 1}$,$W_r\in\mathbb{R}^{512\times 512}$, $b_r\in\mathbb{R}^{512\times 1}$. The activation function $\sigma$ is sigmoid which can be used to normalized the value in [0,1]. And then we can get the combined representation $h$. The dimension of $h$ is $n_0=n+512$.

\begin{equation}
h=\begin{bmatrix} \tilde{\mathscr{L}} \\ \tilde{\mathscr{R}} \end{bmatrix}
\end{equation}


\noindent Then we take $\tilde{h}$ as the input of the multiple layer perceptron(MLP) shown as Eq.~\ref{eq4}.

\begin{equation}
h_1 = \sigma(W_1\tilde{h} + b_1)
\label{eq4}
\end{equation}

\begin{equation}
h_2 = \sigma(W_2h_1 + b_2)
\end{equation}

\begin{equation}
O = W_3h_2
\end{equation}

\noindent Where $W_1\in\mathbb{R}^{n_1\times n_0}$, $b_1\in\mathbb{R}^{n_1\times 1}$ . $W_2\in\mathbb{R}^{n_2\times n_1}$, $b_2\in\mathbb{R}^{n_2\times 1}$, $W_3\in\mathbb{R}^{n_3\times n_2}$. $n_1,n_2$ are dimensions of hidden layers in MLP, $n_3$ is the number of the character classes.In the following, $O$ is fed into softmax classifier.

\section{Model Learning}

The training algorithm is crucial for the system performance of OLHCC, especially for the deep model. Hinton et.al~\cite{hinton_2006} proposed the procedure of two stage strategy to train the deep boltzmann machine~\cite{hinton_2006}. The first stage is to pre-train the deep boltzmann machine layer-by-layer by using the large scale unlabeled data and the second stage is to train the multiple layer perceptron(MLP) ~\cite{mlp} with softmax and fine-tune other boltzmann layers by using the small scale labeled data. In another words, the first stage is unspervised learning and the second stage is supervised learning. This strategy can avoid the gradient diffusion and the lack of the labeled data efficiently. In our architecture, all of our data are labeled data and the scale of the data is fine, although it is far from enough to cover the variation of the OLHCC. The training of the overall framework is divided into two phases, which is analogy to the strategy of the work~\cite{hinton_2006}. In contrast to the ~\cite{hinton_2006}, the two stage of our algorithm are supervised learning. Firstly we train the overall framework, and then we fixed the architecture of the DCNN to train the following layers further. The parameter $W$ of the overall model include two parts:

\begin{equation}
\theta = [\theta_1,\theta_2]
\label{eq5}
\end{equation}

\noindent Where $\theta_1$ is the parameter for the DCNN and $\theta_2=(W_l,b_l,W_r,b_r,W_1,b_1,W_2,b_2,W_3)$ is the parameter for other parts of SSDCNN. The  output of the multiple perceptron layer $O=(o_1,o_2,...,o_{n_3})$ contains the scores for every classes. To obtain the conditional probability $P(y_i|\mathscr{C},\theta)$ for each class $y_i$, we use the softmax operation over all the calsses:

\begin{equation}
P(y_i|\mathscr{C},\theta) = \frac{e^{o_i}}{\sum_{k=1}^{n_3}{e^{o_k}}}
\end{equation}

\begin{Algorithm}[t]{10cm}
\small
\begin{algorithmic}[1]
\Procedure {TRAINING}{$\mathcal{T}$, $W$}
\State {Random initialize $\theta_1$, $\theta_2$}
\State {Initialize $\eta$}\State {$autocorr = .95$}
\State {$fudge\_factor = 1e-6$}
\State {$ historical\_grad = 0$}
\While {the stop condition is not staisfied} \em{\Comment{Phrase I}}
\State {{$\mathcal{B}$} $\leftarrow$ {\em random\_sample}($\mathcal{T}$)}
\State {$historical\_grad += \frac{\partial J(\theta,\mathcal{B})}{\partial \theta}$}
\State {$\mu=  \left. {\frac{\partial J(\theta,\mathcal{B})}{\partial \theta}} \middle/  {(fudge\_factor + \sqrt{historical\_grad})} \right.$}
\State {$\theta \leftarrow \theta +  \mu\times\eta  \frac{\partial J(\theta,\mathcal{B})}{\partial \theta}$}
\EndWhile
\State {$ historical\_grad = 0$}

\While {the stop condition is not staisfied} \Comment{Phrase II}
\State {{$\mathcal{B}$} $\leftarrow$ {\em random\_sample}($\mathcal{T}$)}
\State {$historical\_grad += \frac{\partial J(\theta,\mathcal{B})}{\partial \theta_2}$}
\State {$\mu=  \left. {\frac{\partial J(\theta,\mathcal{B})}{\partial \theta_2}} \middle/  {(fudge\_factor + \sqrt{historical\_grad})} \right.$}
\State {$\theta_2 \leftarrow \theta_2 +  \mu\times\eta  \frac{\partial J(\theta,\mathcal{B})}{\partial \theta_2}$}
\EndWhile
\EndProcedure
\end{algorithmic}
\caption{Training algorithm of SSDCNN. Here $\mathcal{T}$ denotes the training examples, $\mathcal{B}$ is one batch randomly sampled from $\mathcal{T}$, $\theta_1$ the parameter for DCNN, $\theta_2$ the parameter for the mlp with softmax, $\eta$ the initial learning rate in SGD.}
\label{algorithm-curriculum-learning}
\end{Algorithm}
\vspace{-4pt}

\noindent To train our model, the stochastic mini-batch back propagation is used. Given one batch $\mathcal{B}$ of the training instances, the goal of the training algorithm is to minimize the negative log likelihood:

\begin{equation}
J(\theta,\mathcal{B})=-\sum_{k=1}^{|\mathcal{B}|}{logP(y_k|\mathscr{C}_k,\theta)}
\end{equation}

The hyper-parameter learning rate is hard to choose for traditional SGD methods. The adaptive (sub)gradient (AdaGrad) method recently proposed by Duchi~\cite{adagrad} is used for learning rate adaptation. According to the differentiation chain rule and back propagation strategy~\cite{lecun-98b}, for every batch $\mathcal{B}$ the parameter of the architecture can be updated as follows:
\begin{equation}
  \begin{cases}
\theta \leftarrow \theta +  \mu\times\eta  \frac{\partial J(\theta,\mathcal{B})}{\partial \theta}\quad\quad\quad\quad For \quad Phrase \quad  I\\ \\
 \theta_2 \leftarrow \theta_2 +  \mu\times\eta  \frac{\partial J(\theta,\mathcal{B})}{\partial \theta_2}\quad\quad\quad For \quad Phrase \quad  II \\
   \end{cases}
\end{equation}

\noindent Where $\mu$ is calculated by using adaptive (sub)gradient (AdaGrad) method which is updated for every batch. $\eta$ is the given hyper-parameter learning rate. The detailed description of the training algorithm is depicted in Alg.~\ref{algorithm-curriculum-learning}

\section{Experiments}
\subsection{Data Set and Preprocessing}
\begin{figure}
  \centering
  \includegraphics[width=1.0\textwidth]{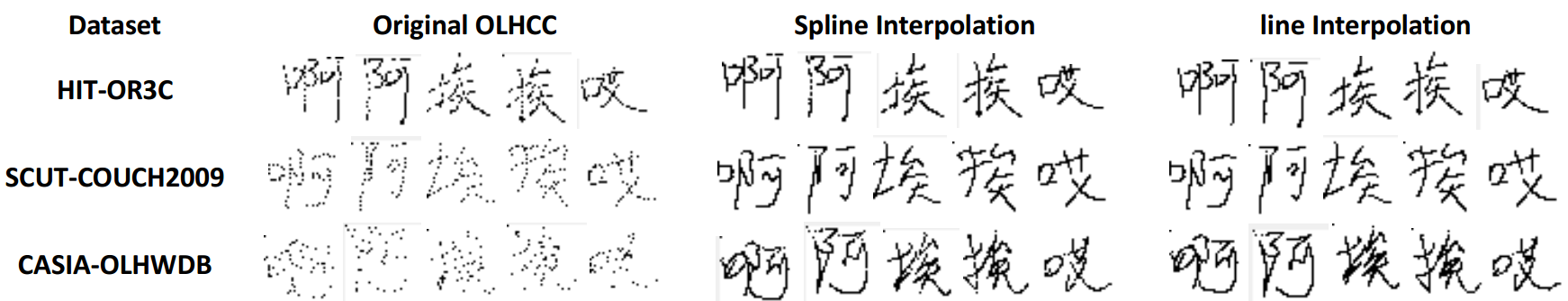}
  \caption{Some OLHCCs from different datasets. The left part shows the original OLHCC from different datasets. The middle part shows OLHCCs after spline interpolation. The right part shows OLHCCs after linear interpolation. }\label{fig5}
\end{figure}

The testbed of our model is the Online Isolated Character Recognition task (Task 3) on ICDAR 2013 Chinese Handwriting Recognition Competition. The organizer provides the recommending training dataset  CASIA-OLHWDB1.0-1.2 and free to use any training dataset~\cite{yinfei2013}. The test set is written by 60 writers who did not contribute to the training dataset~\cite{yinfei2013}. Hence the size of the test set is 60*3755. The character set is confined to the 3,755 Chinese characters in the level-1 set of GB2312-80. In our experiment settings we use the  CASIA-OLHWDB1.1\&1.0 dataset which contain 420 and 300 samples for each character respectively. We also use the  HIT-OR3C~\cite{or3c} and SCUT-COUCH2009~\cite{scut} which contain 280 and 185 samples for each character respectively. Because three types of datasets are constructed by different organizations and the devices they used to collect the dataset are different, which leads to the great difference on the data consistency. Fig.~\ref{fig5} shows some sample characters from three datasets. From top row to bottom, we can see that the number of points in different dataset differs greatly. Some characters even cannot be recognized by human, such as these in third row. In order to reduce data inconsistency, we preprocess the data with interpolation. We tried spline interpolation and linear interpolation.

\subsection{Implementation Detail}

In this part, we will describe the implementation details of our proposed model SSDCNN and some models constructed for the convenience of showing the characteristics of our model. All these models used the adaptive(sub) gradient(AdaGrad) to tune the learning rate and the batch size is 100, the initial learning rate is 0.01. The hyper-parameter selection of deep neural network is difficult, especially when the scale of training dataset is big. In order to select the hyper-parameter efficiently, we choose a subset from the whole training data. The size of this data is 60*3755 OLHCCs, which is reasonable to train an efficient architecture. We use this dataset to train all the models and validate these models on the whole test set. In this way, we choose the hyper-parameters of the final architecture of our models. Table 1 shows the comparison of no interpolation, spline interpolation and linear interpolation in different networks. From the table, we can see that interpolation operations help improving the recognition accuracy greatly compared to no interpolation. For the DCNN architecture, the performances of linear interpolation and spline interpolation are not significantly different. Due to the simplicity of linear interpolation, we adopt linear interpolation. Table \ref{tabl1} shows that the network with four convolutional-pooling layers shows better performance than three layers and five layers. Hence, we select the network with four convolutional and max-pooling layers as our SSDCNN architecture. For eight-direction feature, we first add virtual points when pen is moving from one stroke to another, then normalize the new characters with P2DMN to convert coordinates into fixed-size area. Next, we extract the direction information from the points. After sampling and Gaussian filter from each direction, we obtain feature vectors with 512 dimensions.

\begin{table}[t]
\begin{center}
\begin{tabular}{|c|c|c|c|}
  \hline
  \# of Conv & \# pooling & Interpolation & P@1(\%)
  \\
  \hline
  3 & 3& no &	66.01\\ \hline
  3 & 3 &spline&	80.19\\ \hline
  3 & 3&	linear&	\textbf{80.29}\\ \hline
  4 & 4 & no&	67.21 \\ \hline
  4 & 4	& spline&	81.01\\ \hline
  4 & 4&	linear&	\textbf{81.65}\\ \hline
  5 & 3 & no & 67.05 \\ \hline
  5 & 3 & spline & 80.78 \\ \hline
  5 & 3 & linear & 81.52\\ \hline
\end{tabular}
\caption{Comparison of no interpolation, spline interpolation and linear interpolation in different networks. Each line describes one interpolation operation and its corresponding accuracy in the network. }
\label{tabl1}
\end{center}
\end{table}

Our competitor models include IMDCNN, SSDCNN-8, NN8 which are described as follows. Our competitor models also include all the systems that participate the Online Isolated Chinese Character Recognition task. They are \textbf{HIT}, \textbf{Faybee}, \textbf{TUAT},  \textbf{UWarwick},  \textbf{VO-1},  \textbf{VO-2},  \textbf{VO-3},  \textbf{USTC-1} and \textbf{USTC-2}. The details of these systems are described by Yin et.al~\cite{yinfei2013}.

\begin{itemize}
\item \textsc{IMDCNN:} This is the baseline model which use the static image as input to the DCNN (IMDCNN). The motivation of this model is to show the impact of the stroke sequence on the performance of OLHCCR. The OLHCCs are transformed into bitmaps after line interpolation. The static images are normalized to 32*32.  The positions that the strokes go through are set to ``1'', otherwise ``0''. The architecture of this model can be depicted as: 32*32 -100C3ReLU -MP2 -100C2ReLU -MP2 -100C2ReLU -MP2 -200C2ReLU -MP2 -N100Sig -N3755. In this formula, The substring ``-'' represents a network layer and the ``32*32'' is the size of the input. ``-100C3ReLU'' denotes one convolutional layer where the output feature maps is 100, convolutional window size is $3\times3$, and the activation function is ReLU. ``-MP2'' denotes a max pooling layer, where ``2'' means the size of pooling window is $2\times2$. ``-N200Sig'' denotes one fully-connected layer, where ``200'' is the number of the output nodes and ``sig'' means the activation function is ``sigmoid''. \vspace{-3pt}
\item \textsc{SSDCNN-8:} This model only use the DCNN architecture to learn the stroke sequence-dependent representation without eight-directional features. On one hand, this model can show the ability of stroke sequence-dependent representation. On other hand, it can show the necessity for combining it with eight-directional features. The max number of the stroke are set to 28. Every stroke is set to 32*32. If the stroke number of the character is less than 28. We use the all zero 32*32 paddings. Hence, the input of the DCCN is 28*32*32. The architecture of the network is:28*32*32 -100C3ReLU -MP2 -100C2ReLU -MP2 -100C2ReLU -MP2 -200C2ReLU -MP2 -N100Sig -N3755. \vspace{-3pt}
\item \textsc{NN8:} This model only take the 512 dimension eight-directional features as the input to the MLP. The motivation of this baseline model is to show the ability of the eight-directional features and the necessity for combining it with stroke sequence-dependent representation learned via DCNN. The architecture of the networks is:512 -N300Sig -N200Sig -N3755. \vspace{-3pt}
\item \textsc{SSDCNN:} This is our proposed model which combines the eight-directional features and stroke sequence-dependent representation. This architecture includes three part: DCNN for learning stroke sequence-dependent representation: 28*32*32 -100C3ReLU -MP2 -100C2ReLU -MP2 -100C2ReLU -MP2 -200C2ReLU -MP2 -N200Sig; the eight-directional features:512-N512Sig and the architecture of combining them: 712 -N300Sig -N200Sig -N3755. The DCNN based model needs large scale data to train. However, collecting enough training data is very time consuming. We expand our training dataset from existing training data by randomly drop some points on the stroke. In this way, we can generate some new instances by adding some noises to the existing data. We randomly choose 60*3755 characters from CASIA-OLHWDB1.0-1.1 to expand our training dataset. \vspace{-3pt}
\end{itemize}

\subsection{Experiment Result}
For evaluation, we compare the result of recognition systems with the ground truth to judge whether the result is correct or not. On the test set, we use the precision criteria, which can be described as Eq.~\ref{eval} :
\begin{equation}
P=\frac{N_C}{N_T}
\label{eval}
\end{equation}
Where \(N_C\) is the number of correctly recognized samples, and \(N_T\) is the total number of test samples. In our experiment, The top-rank correct rate and the accumulated correct rate of top 10 classes are often used to evaluate OLHCCR methods.

\begin{table}[!h]
\small
\begin{center}
\begin{tabular}{|m{1.8cm}|m{1.0cm}|m{1.0cm}|m{1.0cm}|m{1.0cm}|}
\hline
\bf{Model} & \bf{P@1} &\bf{P@2}&\bf{P@3}&  \bf{P@10} \\ \hline

Human &95.19&---&---&  ---    \\\hline
IMDCNN & 79.36  &88.07&91.19&  96.51  \\\hline
SSDCNN-8 & 88.01  &93.64&95.60&  98.41    \\ \hline
NN8 &94.87&97.74&98.43&  99.29    \\\hline
\bf{SSDCNN} &\bf{97.44}& \bf{99.33}& \bf{99.72} & \bf{99.86}   \\\hline \hline
RPCNN & 97.39 &--- &---& 99.88  \\ \hline
VO-3 & 96.87  &---&---&  99.67  \\ \hline
VO-2 & 96.72  &---&---&  99.61  \\ \hline
VO-1 & 96.33  &---&---&  99.61  \\ \hline
HIT & 95.18  &---&---&  99.39  \\ \hline
USTC-2 & 94.59  &---&---&  99.14  \\ \hline
USTC-1 & 94.25  &---&---&  99.06  \\ \hline
TUAT & 93.85  &---&---&  99.24 \\ \hline
Faybee & 92.97  &---&---&  98.87 \\
\hline
\end{tabular}
\end{center}
\caption{The performance of different Methods.}
\label{result}
\end{table}

\begin{figure}
  \centering
  \includegraphics[width=.9\textwidth]{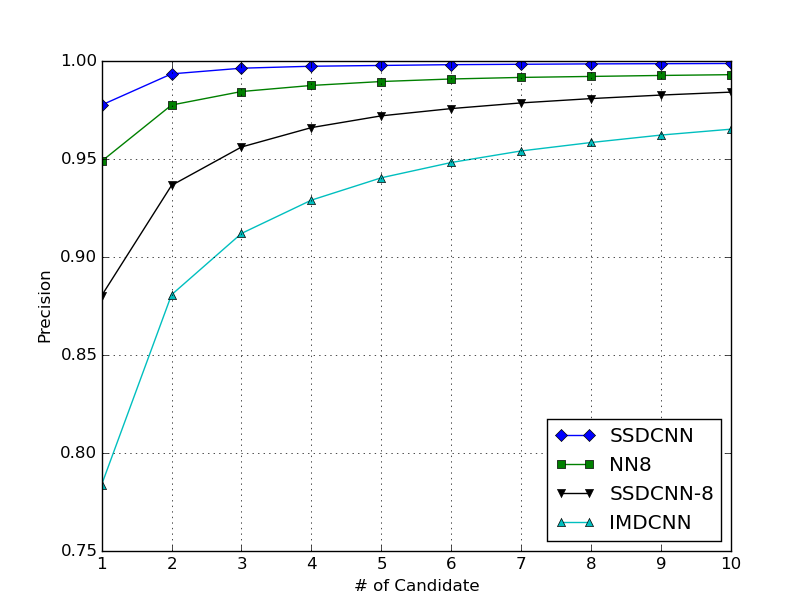}
  \caption{Precision vs different number of candidates.
  }
  \label{candiate}
\end{figure}

The results are listed as Tabel.~\ref{result}. The performance of our SSDCNN is state of art compared to the submitted systems on ICDAR2013 and also outperforms the human performance with a large margin. Tabel.~\ref{result} and Fig.~\ref{candiate} show that most of correct characters of SSDCNN rank in the front(the P@2 is 99.33), which is very important for handwritten Chinese character input method
.

\begin{figure}
  \centering
  \includegraphics[width=.9\textwidth]{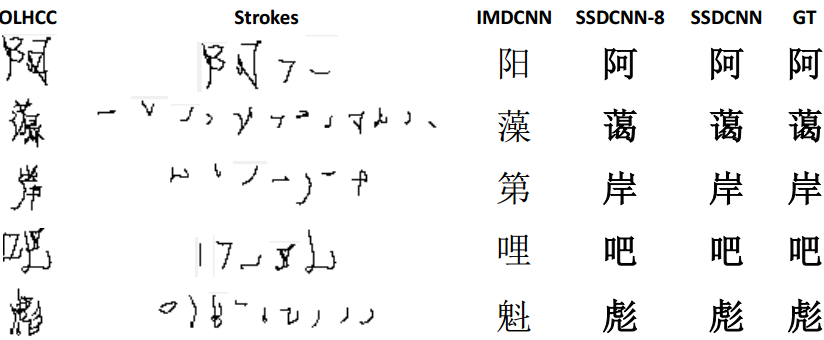}
  \caption{Some Online Handwritten Chinese characters recognized by IMDCNN, SSDCNN-8 and SSDCNN, where results of IMDCNN are wrong. The second column depicts stroke sequence of OLHCCs. The last column is the ground truth.}
  \label{sample1}
\end{figure}

\begin{figure}
  \centering
  \includegraphics[width=.9\textwidth]{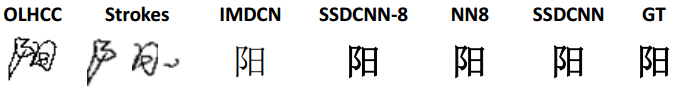}
  \caption{One Online handwritten Chinese character. The ``GT'' column shows the ground truth.}
  \label{yang}
\end{figure}

These results show that SSDCNN-8 outperforms IMDCNN with a large margin (88.01 vs 79.36). This result indicates that the stroke sequence-dependent representation is very important to improve the performance of the DCNN based architecture. The improvement may be attributed to two reasons: 1) the stroke sequence-dependent representation is very effective for some similar handwritten characters. 2) the stroke sequence are beneficial to DCNN for learning the high quality representation of the handwritten characters. Fig.~\ref{sample1} shows some examples SSDCNN-8 can recognize correctly, while IMDCNN failed to recognize them. From these examples, we can see that some OLHCCs are very hard to recognize if the stroke sequence is ignored. For example, in the first row of Fig.~\ref{sample1}, the OLHCC is converted to the static image, it is very like to the OLHCC of Fig.~\ref{yang}. Because of the neglect of the stroke sequence information, the IMDCNN, recognize both of them as \begin{CJK}{UTF8}{gkai}  ``阳'' \end{CJK}. However, the OLHCC in the first row of Fig.~\ref{sample1} is \begin{CJK}{UTF8}{gkai} `` 阿'' \end{CJK}. From the sequence of strokes, we can see that \begin{CJK}{UTF8}{gkai} `` 阳'' \end{CJK} and \begin{CJK}{UTF8}{gkai} `` 阿'' \end{CJK} are  different. Benefiting from the stroke sequence information, SSDCNN recognize \begin{CJK}{UTF8}{gkai} `` 阿'' \end{CJK} and \begin{CJK}{UTF8}{gkai} ``阳'' \end{CJK} correctly, while IMDCNN fails to recognize \begin{CJK}{UTF8}{gkai} `` 阿'' \end{CJK}. These examples indicate that the SSDCCN-8 can exploit stroke sequence information to learn the high-quality representation, which is very helpful for distinguishing confusing OLHCCs.

\begin{figure}
  \centering
  \includegraphics[width=.8\textwidth]{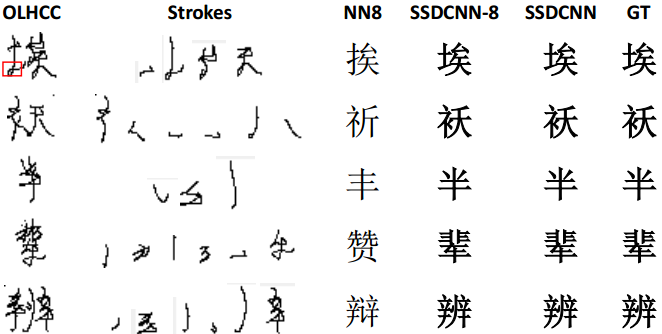}
  \caption{Some Online Handwritten Chinese characters recognized by NN8, SSDCNN-8 and SSDCNN, where results of NN8 are wrong. The second column depicts the stroke sequence of OLHCCs. The last column is the ground truth.}
  \label{sample2}
\end{figure}

\begin{figure}
  \centering
  \includegraphics[width=.8\textwidth]{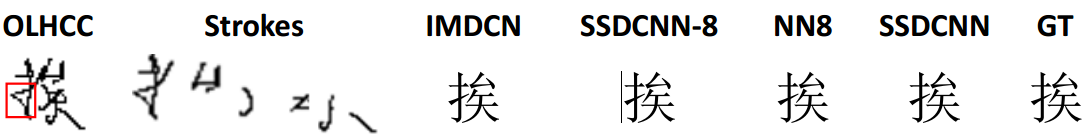}
  \caption{One Online handwritten Chinese character. The ``GT'' column shows the ground truth.}
  \label{ai}
\end{figure}

Even though the NN8 outperforms SSDCNN-8, Fig.~\ref{sample2} shows that some OLHCCs are recognized correctly by SSDCNN while NN8 recognizes them wrongly. Some parts of these characters are very easy to be confused with part of other characters. For example, the left part of handwritten \begin{CJK}{UTF8}{gkai}  `` 挨'' \end{CJK} (shown in Fig.~\ref{ai}) and  \begin{CJK}{UTF8}{gkai} `` 埃 '' \end{CJK}(shown in Fig.~\ref{sample2}) are quite similar. The major difference of them is shown in the red box of Fig.~\ref{sample2} and Fig.~\ref{ai}. However, it is hard for NN8 to learn this slight difference of them by using eight-directional features, because eight-directional features only represents the stroke direction feature by analyzing local points' directions. However, this slight difference is the key feature to distinguish handwritten \begin{CJK}{UTF8}{gkai} `` 挨 '' \end{CJK} and \begin{CJK}{UTF8}{gkai} `` 埃 '' \end{CJK}. The SSDCNN-8 can learn their difference by not only using the stroke sequence information but also considering the global spatial information of points.

\begin{figure}
  \centering
  \includegraphics[width=1.0\textwidth]{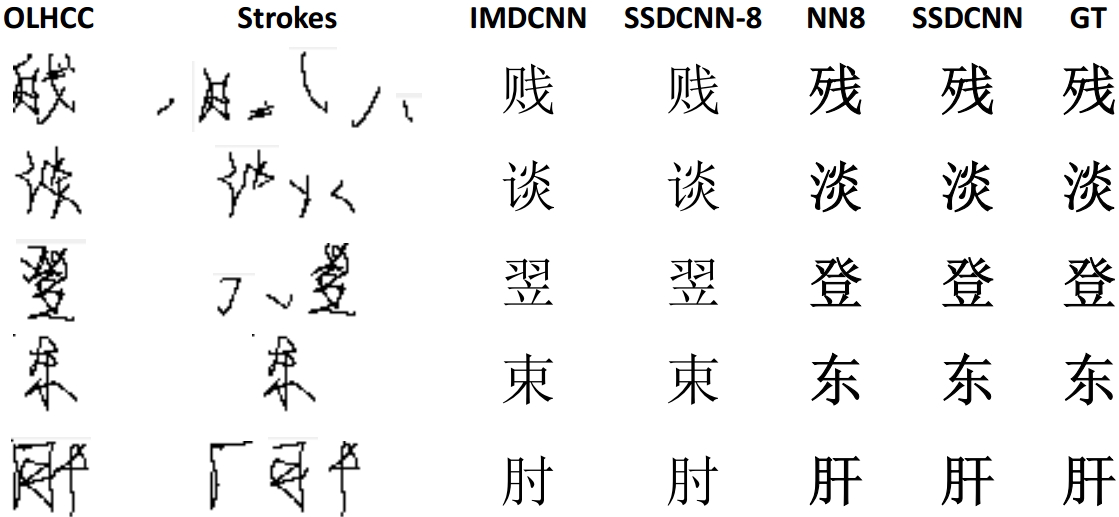}
  \caption{Some Online Handwritten characters recognized by IMDCNN, SSDCNN-8, NN8, where results of IMDCNN and SSDCNN-8 are wrong. The second column depicts the stroke sequence of OLHCCs. The last column is the ground truth.}
  \label{sample3}
\end{figure}

The performance of the NN8 outperforms the SSDCNN-8 and IMDCNN significantly. This result shows that the stroke direction features are very effective. Fig.~\ref{sample3} shows some examples which can be recognized correctly by NN8 and wrongly recognized by IMDCNN and SSDCNN-8. From these examples, we can see that the ground truth of the wrongly recognized OLHCCs by SSDCNN-8 are made up with number of strokes, while the corresponding OLHCCs are represented by less handwritten strokes because of the serious cursive writing. For example, although \begin{CJK}{UTF8}{gkai} ``东'' \end{CJK} is made up of five strokes, the corresponding handwritten character is written by using one cursive stroke. SSDCNN-8 is hard to learn the stroke sequence information for such OLHCCs. Besides, SSDCNN-8 needs large scale data to learn the representation of the OLHCCs because of the scale of the SSDCNN-8 is big and the huge writing variations of characters. The training data may be not enough to sufficiently train SSDCNN-8. Although the eight-directional features can reach close to the human performance, it is difficult to win human only depending on eight-directional features. According to Fig.~\ref{sample1} and Fig.~\ref{sample3}, NN8 is good at recognizing OLHCCs with strokes written clearly and is not sensitive to the number of written strokes, such that \begin{CJK}{UTF8}{gkai} ``东'' \end{CJK} can be recognized correctly. On the other hand, SSDCNN-8 is good at recognizing characters with more handwritten strokes and is less sensitive to the quality of strokes. For example, the \begin{CJK}{UTF8}{gkai} ``蔼'' \end{CJK} in Fig~\ref{sample1} is very hard to recognize by its static image. However, it is written by number of strokes and is recognized correctly by SSDCNN-8. This difference between SSDCNN-8 and NN8 is shown more obviously by the first row in Fig.~\ref{sample3}, SSDCNN-8 recognizes it as character \begin{CJK}{UTF8}{gkai}``贱''\end{CJK}. The left part of OLHCC, is written by only two strokes, there are not enough stroke sequence information can be exploited for SSDCNN-8. Hence, SSDCNN-8 recognizes this part wrongly. However, its right part, \begin{CJK}{UTF8}{gkai} ``戋'' \end{CJK}, contains more handwritten strokes and is recognized correctly by SSDCNN-8.

\begin{figure}
  \centering
  \includegraphics[width=1.0\textwidth]{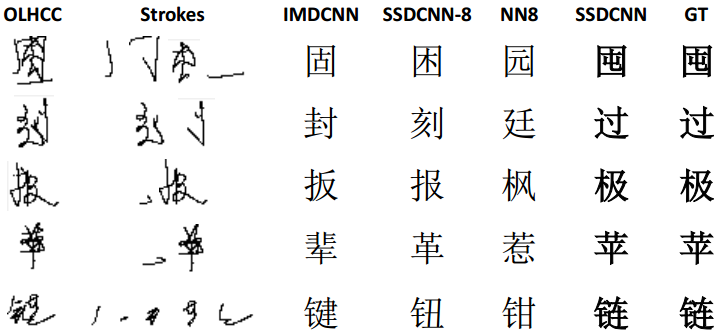}
  \caption{Some Oline Handwritten characters recognized by IMDCNN, SSDCNN-8, NN8 and SSDCNN, where only results of SSDCNN are right. The second column depicts the stroke sequence of OLHCCs. The last column is the ground truth.}
  \label{sample4}
\end{figure}

Since the different advantages SSDCNN-8 and NN8, it is expected that the combination of these two models, i.e., the SSDCNN model, may reach higher performance than each of them. Fig.~\ref{sample4} shows some examples recognized by SSDCNN correctly and wrongly recognized by NN8, IMDCNN and SSDCNN-8. From these examples, we can see that the cursive writing of these characters are very serious. The character \begin{CJK}{UTF8}{gkai} ``苹'' \end{CJK} contains only two cursive strokes, though it's structure is very complex. The two strokes of \begin{CJK}{UTF8}{gkai} ``苹'' \end{CJK} are also not very clear, especially the second stroke. Neither Eight nor SSDCNN-8 can recognize them correctly. When combing them together with SSDCNN, we can distinguish them from their similar characters. Finally, our test result shows that the SSDCNN model that combines stroke sequence-dependent representation with eight-directional features achieve the precision 97.44. This result reduce the error rate by about 50\% compared with NN8. One one hand, this performance indicates that the stroke sequence can benefit to the performance of the OLHCCR. On the other hand, the handcraft features and stroke sequence-dependent representation learned via DCNN have complementary advantages.

\section{Conclusion}
In this paper, we propose a novel deep convolutional neural network model for OLHCCR, named stroke sequence-dependent convolutional neural network(SSDCNN). Instead of pursuing more sophisticate handcraft features, the proposed SSDCNN presents a simple and effective way of learning stroke sequence-dependent representation and integrating the widely used eight-directional features into it. Experiments are conducted on the OLHCCR competition task of ICDAR 2013.  The MLP with only eight-directional features (NN8) can achieve the precision of 94.87\%, while only using the stroke sequence-dependent representation learned by using DCNN can achieve 91.5\%. By using these two types of features, SSDCNN can reduce the error rate of NN8 by about 50\% and it achieves the state of art performance 97.44\%, which reduces the error rate of the best system on ICDAR 2013 competition by 1.9\%. Our results analysis also shows the complementary of stroke sequence representation and the traditional eight-directional features. Since the proposed model is clarity and straightforward, which makes it more easier for further optimizing or adaptation for specific application context. Though the high precision reached by this model, it also open a large room for further improvement. Our future research includes: 1) use the SSDCNN model on larger number of character OLHCC classes. 2) use the subparts of strokes rather than the natural pen-down/pen-lift strokes to address the cursive written issues. 3) use the SSDCNN model for online handwritten recognition of more languages.
\section*{Acknowledgement}

This paper is supported in part by grants: NSFCs (National Natural Science Foundation of China) (61473101 and 61272383), Strategic Emerging Industry Development Special Funds of Shenzhen (JCYJ20140417172417105 and JCYJ20140508161040764)

\section*{References}

\bibliography{my}

\end{document}